%% file: main.tex
\documentclass[runningheads]{llncs}

\usepackage{eccv}

\usepackage{eccvabbrv}

\usepackage{graphicx}
\usepackage{booktabs}

\usepackage[accsupp]{axessibility}  

\usepackage{hyperref}

\usepackage{orcidlink}

\begin{document}

\title{ScAle: Attention Head Scaling as a Minimal Adapter for Spatial Reasoning in Vision–Language Models} 

\titlerunning{ScAle: Attention Head Scaling as a Minimal Adapter}

\author{Rahul Chowdhury\inst{1}\orcidlink{0009-0008-1575-6077} \and
 Timothy A Rupprecht\inst{2}\orcidlink{0000-0002-9574-5261} \and
Xuan Shen\inst{3}\orcidlink{0000-0003-4965-7321}\and
Pu Zhao
\inst{1}\thanks{Corresponding Author}\orcidlink{0000-0001-5018-2859}\and
Yanzhi Wang
\inst{1}\orcidlink{0000-0002-3024-7990}}

\authorrunning{R. Chowdhury et al.}

\institute{Northeastern University, Boston, USA \and EmbodyX Inc., USA \and Zhejiang University, China \\ \email{\{chowdhury.rah, p.zhao\}@northeastern.edu}}

\maketitle

\input{sec/abstract}

\input{sec/intro}
\input{sec/relatedworks}

\input{sec/motivation}

\input{sec/methods}

\input{sec/result_text}
\input{sec/conclusion}



\clearpage  


%
%
\bibliographystyle{splncs04}
\bibliography{main}
\end{document}

%% file: sec/abstract.tex
\begin{abstract}
\label{sec:abs}
Spatial reasoning remains a persistent challenge for many vision--language models (VLMs), and improving it typically requires fine-tuning with substantial additional parameters. Our preliminary analysis reveals that rescaling activations in selected transformer layers—without modifying pretrained weights—can significantly influence downstream performance. Motivated by this observation, we propose \textbf{ScAle}, an ultra-lightweight adaptation method that learns a small set of scalar coefficients to modulate last-token attention and MLP activations in a fully frozen backbone. We evaluate our method on the synthetic spatial reasoning benchmark SpatialEval and on real-world VQA datasets (COCOQA and VGQA) across multiple model families. Our method, \textbf{ScAle}, achieves up to \textbf{134.1\% relative accuracy gains} using only \textbf{1K} trainable parameters without requiring millions of trainable parameters as in standard 
LoRA. Despite its extreme compactness, our approach recovers a substantial fraction of standard PEFT performance while preserving strong non-spatial VQA accuracy. These results demonstrate that bounded activation reweighting provides a simple, architecture-agnostic, and highly parameter-efficient alternative for adapting pretrained VLMs. Code is available at \url{https://github.com/rchowdhubnor/ScAle.git}. 
\end{abstract}

  \keywords{Bounded Activation Scaling \and Parameter-Efficient Fine-tuning \and Spatial Reasoning \and Vision–Language Model}

%% file: sec/intro.tex
\section{Introduction}

Spatial reasoning is an important capability for vision--language models (VLMs)~\cite{qwen2.5-VL ,taherin2025cross,zhao2025open,shen2024numerical,shen2024lazydit,liu2023llava}, enabling models to localize, compare, and reason about entities in structured environments. Despite recent progress, benchmarks such as SpatialEval~\cite{wang2024picture} show that even large VLMs struggle with geometric relations, navigation, and map-based reasoning across model scales and families.

In preliminary experiments, we observe that simply rescaling activations 
of certain layers—without modifying pretrained weights—can significantly influence downstream spatial reasoning performance. Specific layer and scalar combinations produce significant shifts in accuracy, indicating that targeted activation scaling can meaningfully reweight internal representations. This observation motivates us to investigate whether learnable activation scaling can serve as a lightweight alternative to conventional parameter-efficient fine-tuning (PEFT).

We propose \textbf{ScAle}, a minimal adaptation mechanism that applies a single learnable scalar to each attention head output at the last token of a frozen transformer backbone. This operation preserves the pretrained computation while enabling controlled amplification or attenuation of task-relevant signals. The scaling function is differentiable and bounded, ensuring numerical stability and compatibility across architectures.

Across Qwen2.5-VL (3B, 7B)~\cite{qwen2.5-VL} and LLaVA-Next (7B, 13B)~\cite{liu2024llavanext} models on SpatialEval, our method achieves up to 130\% relative accuracy gains over frozen baselines using approximately 1K trainable parameters. Compared to LoRA configurations requiring millions of parameters, our approach achieves over three orders of magnitude greater parameter efficiency while recovering a substantial fraction of full PEFT performance.
We summarize our contributions as follows:
\begin{itemize}
\item Our preliminary experiments show that  activation reweighting in one single layer of frozen VLMs can substantially improve spatial reasoning accuracy.
\item We introduce ScAle, an ultra-lightweight adaptation strategy that learns as few as one scalar per head applied at the last token to improve the spatial reasoning capability effectively.
\item We show that our method can achieve consistent gains across multiple model families and data regimes on the synthetic spatial reasoning benchmark SpatialEval and  spatial reasoning benchmark What’sUp, 
while requiring dramatically fewer trainable parameters than baselines. 
\end{itemize}

%% file: sec/relatedworks.tex
\section{Related Work}
\label{sec:related}

\noindent\textbf{Spatial Reasoning in Vision--Language Models.}
Spatial reasoning benchmarks such as SpatialEval~\cite{wang2024picture} reveal that even large VLMs struggle with geometric relations, map understanding, and object localization. Prior works have improved spatial understanding through fine-tuning on specialized dataset with structural and spatial grounding like ~\cite{Chen_2024_CVPR,cheng2024spatialrgpt,ogezi-shi-2025-spare}. These methods, however, often require costly retraining and degrade generalization to other reasoning tasks.


\noindent\textbf{Parameter-Efficient Fine-Tuning.} LLMs and VLMs typically require massive resources for training and inference~\cite{zhao-etal-2024-pruning,10.1145/3418297,zhan-etal-2024-rethinking-token,zhan2024exploring,zhang2022advancing,shen2025sparse,shen2025quartdepth,yang2026survey,li2024pruning,kong2025}.
Parameter-efficient fine-tuning (PEFT) techniques adapt pretrained transformers with minimal parameter overhead~\cite{hu2022lora,shen2025draftattention,yang2023pruning,shen2025fastcar,shen2025efficient,shen2024search,zhan2024fast,mi2026effective}. LoRA~\cite{hu2022lora} introduces low-rank adapters into linear projections, while (IA)$^3$~\cite{liu2022few} learns layer-specific vectors of scaling factors to modulate every dimension of key, value, and feed-forward activations. Prefix-tuning~\cite{li-liang-2021-prefix} learns a small set of vectors prepended as to the input. While \cite{houlsby2019parameter} introduces adapter-tuning, where small task-specific networks are injected and trained within each transformer layer, these modules still add a substantial number of parameters compared to our minimalist design. These methods achieve strong downstream adaptation but still introduce tens of thousands to millions of additional parameters. Our method differs by introducing only a handful of bounded scalars—one per head—that continuously scale existing activations, achieving fine-grained control with far
fewer  parameters.

\noindent\textbf{Activation Modulation and Interpretability.}
Activation-level interventions have gained prominence in mechanistic interpretability and model editing. Studies have shown that scaling or patching activations can expose causal circuits within transformers~\cite{vig2020investigating,meng2022locating,wang2022interpretability,goldowsky2023localizing}. The work~\cite{hua2025visionlanguagemodelsprocessconflicting} demonstrates that scaling attention heads can reveal modality-specific pathways in multi-modal transformers. Our approach extends these ideas by learning such scaling coefficients end-to-end, producing interpretable, differentiable modulation of the residual stream that maintains numerical stability through bounded activation scaling. Popular activation steering techniques  \cite{li2023inference,turner2023steering,rimsky2024steering} construct steering vectors and use them to adjust residual stream to obtain target behavior without retraining or fine-tuning. While SteerVLM~\cite{sivakumar2025steervlm} extends activation steering into the multimodal domain by introducing a parameterized Steering Module that learns nonlinear adjustments from paired target and converse prompts, it requires to finetune millions of parameters (14\% of model parameters) with high training cost. Different from previous works, our method only needs to finetune  hundreds to thousands of parameters, which is lightweight and cost-saving.

%% file: sec/motivation.tex
\section{Background and Motivation}



Although VLMs achieve superior performance in various downstream tasks, recent benchmarks such as SpatialEval~\cite{wang2024picture} reveal that even the most capable VLMs struggle with spatial reasoning, such as geometric relations, map understanding, and directional inference. 
These deficiencies persist across scales and modalities, underscoring a broader limitation of spatial reasoning capabilities.

\subsection{Spatial Reasoning Tasks}

\begin{figure*}[t]
\centering
\includegraphics[width=0.9\textwidth]{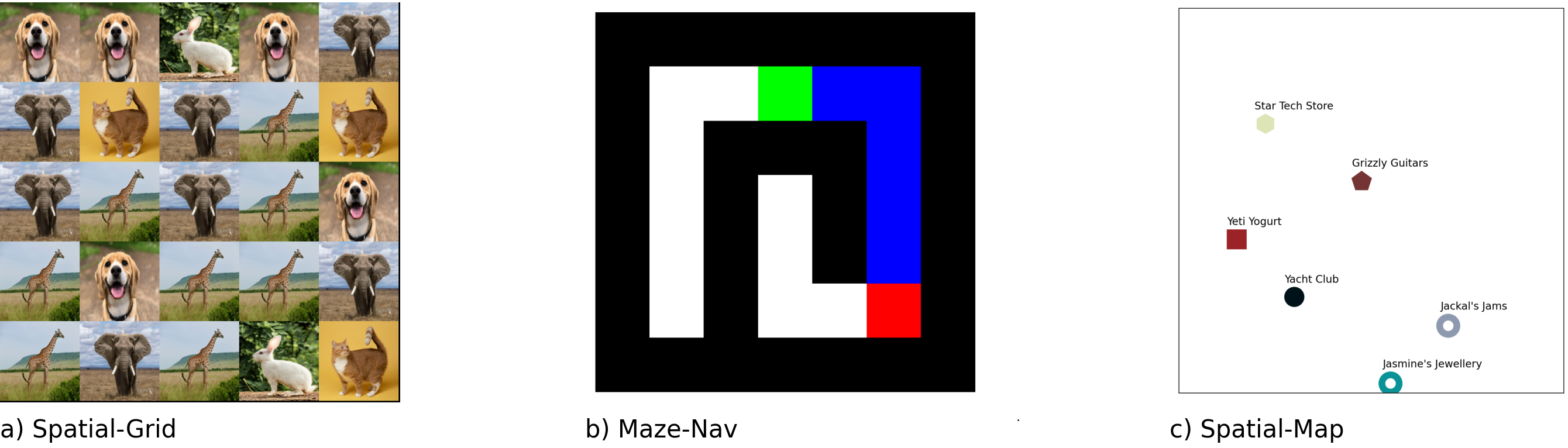}
\caption{Examples  of SpatialEval: (a) Spatial-Grid, (b) Maze-Nav, and (c) Spatial-Map. 
}
\label{fig:Dataset}
\end{figure*}

The SpatialEval benchmark~\cite{wang2024picture} is designed to evaluate the spatial reasoning capacity of LLMs and VLMs.  
SpatialEval includes multiple sub-tasks  to cover various  critical aspects of spatial reasoning (with examples shown in Fig.~\ref{fig:Dataset}): 
\begin{itemize}
\item \textbf{Spatial-Grid} reveals how well a model can perform coordinate-based  positional reasoning with explicit spatial regularity   in structured environments.
\item \textbf{Maze-Nav}, targets navigation properties such as counting turns, identifying right/left turns, or determining start and exit locations. 
Maze-Nav challenges the model to understand  orientation changes, integrate sequential spatial information and reason about translations in two-dimension. 
\item \textbf{Spatial-Map}, assesses  if models can form a coherent internal representation of spatial relations between named entities  across a map-like plane.
\end{itemize}
Together, these three tasks cover critical aspects of spatial reasoning: Spatial-Grid emphasizes identification of the location target object in the scene, Maze-Nav examines sequential spatial inference, and Spatial-Map measures spatial relationships between objects in the scene. This benchmark with a broad spectrum of spatial reasoning tasks  provides a rigorous testbed for our evaluation.

\subsection{Failure of VLMs on Spatial Reasoning}

Although VLMs are powerful, their spatial reasoning capabilities may be limited. 
We evaluate the performance of  \textbf{LLaVA-Next}~\cite{liu2024llavanext}, the successor to \cite{liu2023improvedllava} and \cite{liu2023llava}, on  SpatialEval~\cite{wang2024picture}  with three tasks: Spatial-Map, Maze-Nav, and Spatial-Grid. 
As  observed in Fig.~\ref{fig:manual scaling}, the original model with an unaltered forward pass  (outlined by the  vertical band in black with a scaling factor of $1.0$), achieves low accuracy (around 20\%$\sim$30\%) on different spatial reasoning tasks. It shows that although VLMs are superior, their spatial reasoning capabilities may be limited.


\subsection{Spatial Calibration Hypothesis}
We hypothesize that the limited spatial reasoning performance is due to the mode competing in LLMs.  Specifically, LLMs maintain multiple competing modes.
For spatial reasoning tasks, spatial cues are encoded in a redundant, weakly ordered 1-D token stream.  Its  2-D spatial reasoning  signal can be obscured by interference from other activated modes. 
Our objective is to disambiguate these competing modes and steer the model toward the mode that is most appropriate for the specific spatial reasoning task at hand.
To achieve this, we use activation-level adjustments to scale the activations within a bound, thus activating/promoting the spatially specialized modes without altering any pretrained weights.



\subsection{Preliminary Observations}
\label{sec:manual scaling}


\begin{figure*}[t]
\centering
\includegraphics[width=0.9\textwidth]{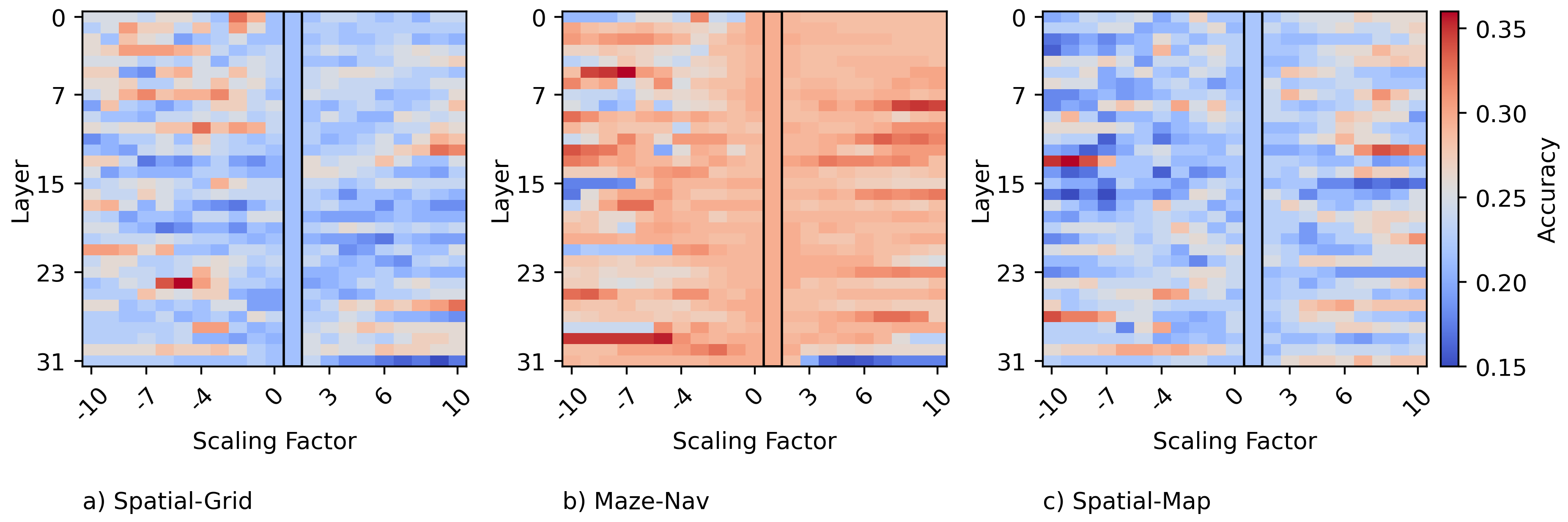} 

\caption{
Accuracy heatmaps for multiplying one scalar to the last-token MLP activations in one single layer. Different scalars ranging from $-10$ to $+10$ are applied to different layers to obtain the heatmap. 
We evaluate LLaVA-Next 
on three SpatialEval tasks. 
}

\label{fig:manual scaling}
\end{figure*}

Based on our model competing hypothesis, we explore whether the spatial reasoning performance can be improved with a \emph{tiny, bounded set of scalars} that rescale the final text token inside each transformer layer’s MLP block. 
In detail, we apply activation scaling to the last token of the MLP outputs in one transformer layer for the \textbf{Vicuna-7B} backbone in \textbf{LLaVA-Next}~\cite{liu2024llavanext}. 
We evaluate this modulation on the SpatialEval~\cite{wang2024picture} benchmark, which comprises three tasks: {Spatial-Map}, {Maze-Nav}, and {Spatial-Grid}. 
The scaling factors are swept from $-10$ to $+10$ during the forward pass, with $1.0$ denotes the original  forward without scaling  activations. 
All pretrained model parameters are \textbf{frozen}.

Fig.~\ref{fig:manual scaling} shows the accuracy when scaling different layers with different scaling factors. We can observe that, by scaling MLP last-token activations  of only one single layer, the accuracy can be improved significantly denoted by the emerging red hotspots, compared with the unscaled baseline (with a scaling factor of $1$).

This performance boost can arise from the influence of scaling on downstream computations, where suppressive activations are attenuated and task-relevant pathways are amplified, thereby promoting modes within the network that specialize in spatial reasoning. 
This motivates us to learn a set of scaling factors, rather than relying on the brute-force search on a single MLP or attention head.

%% file: sec/methods.tex
\section{Methods}
Inspired by our preliminary findings, we introduce a simple yet effective approach that learns a set of scalar coefficients to modulate internal activations while keeping all pretrained parameters frozen. We deliberately choose to \emph{learn} these scalars, 
since pretrained transformers exhibit smooth, well-structured embedding spaces. Backpropagating through the model enables different layers to coordinate and calibrate these modulation strengths. 
We apply the learned scaling factors only at the last token. 
Sharing these scalars across all tokens would entangle contextual encoding with query specific adaptation and could disrupt upstream representations. Our objective is to achieve PEFT by learning scaling factors that recalibrate internal representations for spatial reasoning. 

\subsection{Problem Formulation}
We study whether a frozen vision--language model (VLM) can be steered toward spatial multiple-choice reasoning without updating its backbone parameters. Each example from the SpatialEval benchmark~\cite{wang2024picture} consists of an image $  I$, a natural-language question $  q$, and four candidate answers $\mathcal{C}= \{  A,  B,  C,  D\}$, with one correct option $  c^\star \in \mathcal{C}$. Let $f_{  \theta}(  c^\star |   I,   q)$ denote the probability of choosing the correct answer $  c^\star$ from a pretrained VLM with its original parameters $  \theta$, given the image $  I$ and the question $  q$. 
At inference time, the model produces token-level logits, 
and we only use the final token to score the four answer options. 

For each task, we randomly generate a train/test split at the beginning and keep it fixed across all experiments to ensure consistency. To conduct a comprehensive study, we vary the number of training samples and evaluate three regimes with 75, 150, and 300 examples, corresponding to 5\%, 10\%, and 20\% of the 1500 total samples per task. The remaining 95\%, 90\%, and 80\% of samples are used for testing, respectively. During training, for each task, the objective is to learn a small set of \emph{bounded scaling parameters} $  \phi$ that modulate internal activations at the last token. After scaling the internal activations, the new probability of correct answer can  be denoted by $f_{  \theta,  \phi}(  c^\star |   I,   q)$. Our training target is to maximize this probability by learning the scaling parameters $\phi$ as below, 
\begin{align}
\small
\hat \phi \;=\; \arg\max_{\phi} f_{\theta,\phi}(  c^{\star} |  I,   q).
\end{align}
Note that the original model parameters $\theta$ are   kept frozen, and all adaptation capacity arises solely from $\phi$.

\subsection{Base Models}
We employ four VLMs  from  two widely used model families (\textbf{Qwen}~\cite{qwen2.5-VL,bai2025qwen25vltechnicalreport,Qwen2VL,Qwen-VL} and \textbf{LLaVA}~\cite{liu2024llavanext,liu2023improvedllava,liu2023llava}) under various parameter scales (3,7 and 13 billion parameters): Qwen2.5-VL-3B-Instruct, Qwen2.5-VL-7B-Instruct, llava-v1.6-vicuna-7b-hf, and llava-v1.6-vicuna-13b-hf. All models couple a visual encoder with a transformer-based language backbone comprising $L$ layers. Our method operates exclusively on the language backbone, where we learn bounded scaling factors to modulate hidden activations without updating any of the pretrained parameters.

\subsection{Bounded Last-Token Scaling}
We 
apply a bounded, learnable scaling factor to the activations of the \emph{last token} at each layer.  
We use a  function to bound the scalars as below,
\begin{align}
\small
\text{eff}(   u) = 1 +    s_{\max}\tanh(   u),
\end{align}
where $u$ is a scalar parameter and $ s_{\max} > 0$ limits the deviation from $1$. This ensures stability by keeping scaling within $(1-   s_{\max}, 1+   s_{\max})$. For all of our experiments, we choose $   s_{\max}$ to be 10.

\subsubsection{Last-token scaling for attention heads.}
For finer control, we allow per-head scaling on    the last token of the attention outputs. If the $l^{th}$ layer  has $  H$ heads with dimension $   d_h$, the last token in the output of each head $   h$ can be denoted by $   z_{l,h}\!\in\!\mathbb{R}^{   B\times    T\times    d_h}$, where $   B$ is the batch size, and $   T$ is the token number. 
Since we  just modulate the last token, the value of $   T$ is $1$.
The  modulated activation of each head can be denoted by 
\begin{align}
\small
\tilde{z}_{l,h} = \text{eff}\!\left(u^{\text{attn}}_{l,h}\right)\cdot z_{l,h}.
\end{align}
where $u^{\text{attn}}_{l,h}$ is the learnable scalar corresponding to the $h^{th}$ head in the $l^{th}$ layer.
After scaling, the modulated activation is then passed to the output projection $W_O$. 
Note that we assign one single scalar to  each head in each layer, the total number of scalars which need to be trained are $L\times H$.

\subsubsection{Last-token scaling for MLPs.} For the MLP in the  $l^{th}$ layer, we maintain one learnable scalar $u^{\text{MLP}}_l$. The last token of the MLP output is denoted by $Y_l \in \mathbb{R}^{B\times T\times d}$,  where $B$ is the batch size, $T$ is the token number, and $d$ is the dimension of MLP activation. Similarly, $T=1$ since we  only modulate the last token. The scaled MLP last token can be represented by
\begin{align}
\small
\tilde{Y}_l = \text{eff}\!\left(u^{\text{MLP}}_l\right)\cdot Y_l.
\end{align}
We assign only one scalar for the whole MLP outputs in each layer. Thus,  the number of learnable parameters for MLP equals to $L$, much smaller than LoRA.

\subsection{Training Objective}
For each spatial reasoning task 
and training size $K$, we use the same fixed disjoint train/test splits to ensure a fair comparison. Given logits $\ell_\phi(I,q)$ for a sample $(I,q,c^\star)$, we minimize the following loss, 
\begin{align}
\small
\mathcal{L}(\phi)
=
\text{CE}\!\left(\ell_\phi(I,q), c^\star\right)
+ \lambda_{\text{attn}}\sum_{l,h} \|u_{l,h}^{\text{attn}}\|_2^2
+ \lambda_{\text{MLP}}\sum_{l} \|u_{l}^{\text{MLP}}\|_2^2
\end{align}
\begin{align}
\small
\phi^\star
=
\arg\min_{\phi}
\mathbb{E}_{(I,q,c^\star)\sim \mathcal{D}_{\text{train}}}
\mathcal{L}(\phi)
\end{align}

The loss is basically a cross-entropy loss with an $\ell_2$ norm regularization. To perform a comprehensive study, we investigate three cases: scaling only attention activations, scaling only MLP outputs, and both.  When only modulating attention activations, $\phi$ only have attention related trainable scalars  $u_{l,h}^{\text{attn}}$ and $\lambda_{\text{MLP}}$ is set to 0. Similarly, for only scaling MLP outputs, $\phi$ only have MLP related trainable scalars  $u_{l}^{\text{MLP}}$ and $\lambda_{\text{attn}}$ is set to 0.

\subsection{Advantages}
\label{subsection:param}
\subsubsection{Superior Parameter Efficiency.}
Our method leads to superior parameter efficiency which  trains significantly less parameters than typical finetuning methods.
Compared with the  standard LoRA~\cite{hu2022lora} with rank 1 on all transformer blocks, our method requires training significantly less parameters,  as shown in Table~\ref{tab:spatialeval_best_acc_upto}, where the full LoRA needs to train above 2500$\times$ more parameters. 
Compared with a much simpler version of LoRA (named LoRA-last) which only applies LoRA to  the last transformer block with rank 1, as shown in Table~\ref{tab:spatialeval_best_acc_upto}, our method still leads to $61\times$ to $115\times$ parameter efficiency with much less trainable parameters. 
The results demonstrate that we can achieve comparable or much better performance  with far fewer trainable parameters than LoRA at the last layer.

Compared with (IA)$^3$~\cite{liu2022few}, our method needs much less trainable parameters. Specifically,  (IA)$^3$~\cite{liu2022few} learns three scaling vectors per layer, assigning one scaling coefficient per activation dimension.
In contrast,  our method learns a single bounded scalar that scales the entire activation vector (e.g., the output of a head or MLP) as a whole, yielding extreme parameter efficiency.  Prefix-tuning~\cite{li-liang-2021-prefix} reduces the adaptation cost but still inserts thousands to millions of weights, obscuring how local parameter changes translate into functional behavior.


\subsubsection{Promoting Spatial Reasoning.}
Hua \etal~\cite{hua2025visionlanguagemodelsprocessconflicting} use activation scaling of attention heads to identify those that promote one modality of evidence over another when conflicting information is present in the input to a VLM. 
In contrast, rather than using activation scaling to locate heads for modality selection in isolation, we learn scaling factors jointly across all layers and attention heads using standard gradient based optimization. Specifically, we apply backpropagation to train per head scaling parameters while keeping the underlying model weights frozen. We position spatial reasoning as a motivating case study to examine whether such global yet lightweight modulation can calibrate a frozen model. Because scaling adjusts activation magnitudes without introducing new representational directions as LoRA~\cite{hu2022lora} does, we do not interpret it as learning new skills, but rather as reweighting and promoting task relevant modes that already exist within the network.

%% file: sec/result_text.tex
\section{Evaluation}

\subsection{Experiment Setup}

We employ four VLMs  across different model families and  parameter scales: Qwen2.5-VL-3B-Instruct, Qwen2.5-VL-7B-Instruct, llava-v1.6-vicuna-7b-hf, and llava-v1.6-vicuna-13b-hf. We compare our method with baselines on various datasets primarily for spatial reasoning including SpatialEval~\cite{wang2024picture} (Maze-Nav, Spatial-Map, Spatial-Grid), and WhatsUp-VLM \cite{kamath2023whatsup} (COCOQA and VGQA). 

\paragraph{Baselines.} We compare with the following baselines: (1) LoRA-all~\cite{hu2022lora}, which applies rank-1 LoRA to the whole model; (2) LoRA-last~\cite{hu2022lora}, which applies rank-1 LoRA to the last layer of the model with fewer trainable parameters; (3) (IA)$^3$~\cite{liu2022few}, which rescales inner activations with learned vectors.

\paragraph{Setup.}  We train the scaling parameters using limited data (with $K$ training data samples) for 5 epochs. To conduct a comprehensive study, 
we vary the value of $K$ for 75, 150, and 300, respectively, corresponding to 5\%, 10\%, and 20\% of the 1500 total samples per task. The remaining 95\%, 90\%, and 80\% samples are used for testing, respectively.
We apply our method to (1) only attention modules, (2) only MLP modules, or  (3) both attention and MLP, respectively.   

\paragraph{Training Hyperparameters.} We update the trainable scalars with  learning rate of $1\mathrm{e}{-3}$ and set   $\lambda_{\text{attn}}$ or  $\lambda_{\text{MLP}}$ to $1\mathrm{e}{-4}$.  AdamW is adopted as the optimizer with micro-batch-size of 1. Since $\theta$ is frozen, the optimization adjusts only the scaling coefficients $\phi$ that modulate residual strength at each layer.

\subsection{Main Results for ScAle}
As shown in Table~\ref{tab:spatialeval_llava7b}, we evaluate ScAle on SpatialEval using LLaVA-7B and observe that the frozen backbone exhibits limited spatial reasoning capability, achieving only 31.7\%, 29.3\%, and 22.7\% accuracy on Maze-Nav, Spatial-Grid, and Spatial-Map, respectively. By only introducing only 1{,}024 trainable attention-scaling parameters, ScAle (attn) substantially improves performance to 68.9\%, 68.6\%, and 45.0\% under the 20\% training split, corresponding to relative gains of 117.3\%, 134.1\%, and 98.2\%, respectively. Remarkably, ScAle approaches the performance of LoRA-all (2.6M parameters) while using over 2{,}500$\times$ fewer trainable parameters, demonstrating that bounded activation scaling directly targets the locus of spatial reasoning.

\input{tables/1spatial_llava7b}

\subsubsection{Superior Parameter Efficiency.}
As shown in Table~\ref{tab:spatialeval_llava7b}, our method can improve the spatial reasoning accuracy significantly with only 1,024 trainable parameters.  
Although LoRA-all attains higher absolute accuracy, it requires approximately 2.65M trainable parameters ($2{,}600\times$ more  than our ScAle), with significantly higher training efforts. Similarly, (IA)$^3$ needs to finetune much more parameters ($600\times$) than our method. LoRA-last performs much worse than our ScAle (attn) even though it finetunes $76\times$ more parameters.

\subsubsection{Effective Scaling for Attention.}
We notice in Table~\ref{tab:spatialeval_llava7b} that, ScAle (attn) and   ScAle (both) achieve similar accuracy, while ScAle (mlp)  does not lead to significant improvements due to too few scalars. Notably, attention scaling consistently outperforms MLP scaling, indicating that spatial reasoning failures primarily arise from mis-calibrated token interactions rather than insufficient feed-forward capacity. Since attention layers govern relational composition across spatial tokens, lightweight modulation at this stage effectively re-balances spatial aggregation without altering pretrained weights. Thus, for simplicity, we mainly report the results of ScAle (attn) in the following sections.

\subsubsection{Generalization Across Various Models.}

To evaluate architectural robustness, we apply ScAle across four vision-language backbones: LLaVA-7B, LLaVA-13B, Qwen2.5-VL-3B, and Qwen2.5-VL-7B, reporting results under the 20\% training split in Table~\ref{tab:spatialeval_best_acc_upto}. By introducing only $\sim$1K trainable parameters (1.0K, 1.6K, 576, and 784 parameters respectively), ScAle consistently yields substantial improvements over frozen baselines across various model families and scales.

\input{tables/2spatial_all}

On Maze-Nav, ScAle improves the accuracy of  LLaVA-13B  from 26.7\% to 69.8\% (+43.1 absolute points) using only 1.6K parameters, and improves Qwen2.5-VL-3B from 26.0\% to 70.3\% (+44.3 points) with just 576 parameters. Even for stronger backbones such as Qwen2.5-VL-7B, ScAle boosts performance from 38.3\% to 66.2\%. Similar trends hold on Spatial-Grid and Spatial-Map, where lightweight scaling consistently closes a large portion of the gap to full LoRA. 

Importantly, these gains are achieved without architecture-specific modifications, indicating that activation scaling functions as a model-agnostic adapter mechanism. While increasing adaptation capacity to several million parameters (e.g., LoRA-all) yields further improvements, the marginal gains are disproportionately small relative to the parameter increase. These results demonstrate that substantial cross-architecture accuracy gains can be unlocked with only $\sim$1K parameters, establishing activation scaling as a robust and lightweight adaptation strategy for spatial reasoning.

\subsection{Performance on Real-World Spatial Reasoning Tasks}
\subsubsection{Real-World Spatial Reasoning Evaluation.}
To demonstrate the general performance,  we apply our method to real-world spatial reasoning tasks with  WhatsUp-VLM  \cite{kamath2023whatsup} that includes VGQA and COCOQA. As shown  in Table~\ref{tab:whatsup_last_epoch_rebuttal}, when trained on 20\% of the data and evaluated on the remaining 80\%, our method improves the original 56\% accuracy to above 90\% for both VGQA and COCOQA on LLaVA-7B.  For Qwen2.5-VL-3B, our method can also improve the accuracy to above 93\% with approximately 10\% accuracy increase.  Our method only needs to additionally finetune 1K parameters or less, while (IA)$^3$ trains $600\times$ more parameters with much higher training cost. 
\input{tables/3whatup}

\subsubsection{Hallucination Robustness.}

We further evaluate hallucination robustness using the POPE benchmark~\cite{li2023evaluating}, which measures a model’s tendency to hallucinate objects that are not present in the image. Unlike our spatial reasoning benchmarks, POPE does not explicitly test spatial relationships. Instead, it evaluates visual grounding and faithfulness—whether the model’s predictions are aligned with actual visual evidence.

This benchmark provides an important complementary testbed: while spatial reasoning evaluates \emph{where and how} objects relate, hallucination robustness evaluates \emph{whether} the predicted objects are visually grounded at all. Both settings probe visual faithfulness, but from different perspectives.

As shown in Table~\ref{tab:whatsup_last_epoch_rebuttal}, trained on 100 randomly selected sample and tested on 1000 samples, our lightweight activation scaling improves or matches hallucination robustness using only $\sim$1K parameters. On LLaVA-7B, ScAle increases POPE accuracy from 82.1\% (frozen) to 88.2\%, closing most of the gap to (IA)$^3$ (89.8\%) while using \textbf{1.02K} parameters compared to 664K. On Qwen2.5-VL-3B, our method improves accuracy from 87.3\% to 90.4\%, matching (IA)$^3$ performance (90.4\%) with only 576 trainable parameters versus 524K.

\input{tables/5pope}

These results indicate that activation scaling not only enhances spatial reasoning, but also improves visual grounding and reduces hallucinations. Importantly, the gains are achieved with orders of magnitude fewer parameters, reinforcing that ScAle operates as a highly parameter-efficient targeted adapter.
\subsection{Cross-Task Evaluation}

To perform a comprehensive study, we conduct a cross-task evaluation for Qwen-3B, where ScAle trained on one task is evaluated on different datasets. 


\input{tables/4cross_task}

As shown in Table~\ref{tab:cross_task_grid}, ScAle trained on spatial tasks transfers strongly from synthetic to real-world spatial reasoning settings, including VGQA and COCOQA. For example, training on spatial-grid improves the accuracy on VGQA from 84.5\% to 92.2\% and COCOQA from 83.9\% to 90.9\%. 
This indicates that the learned activation scaling captures transferable spatial structure rather than task-specific heuristics, thus preserving or improving performance when a ScAle adapter trained on synthetic data is transferred to related real-world tasks.  

Beyond spatial transfer, ScAle largely preserves general language understanding performance. 
As  demonstrated in Table~\ref{tab:cross_task_grid},
Our adapters trained on Spatial-Grid  can improve the accuracy of GLUE/MRPC~\cite{wang2018glue} from 66.7\% to 73.8\%, and Maze-Nav training achieves 74.5\%
On GLUE/SST2~\cite{wang2018glue}, most spatially-trained adapters maintain comparable performance, with Spatial-Map leading to  significant improvement over frozen (88.3\% vs. 65.8\%).
We can observe that the our learned lightweight activation  scaling can enhance spatial reasoning without degrading (or  even improving) general language understanding performance.


Crucially, we do not observe catastrophic degradation across tasks. Improvements on spatial reasoning benchmarks (including VGQA and COCOQA) occur without systematically harming unrelated language benchmarks. This suggests that learned activation scaling functions as a localized and structured adapter, modulating task-relevant directions in representation space rather than globally altering the pretrained model.

Overall, our lightweight activation scaling maintains strong general VQA performance while consistently improving spatial reasoning across both synthetic and real-world benchmarks.


\subsection{Ablation Study}
As shown in Table~\ref{tab:spatialeval_llava7b}, we ablate applying our method to attention blocks, MLP blocks, or both. ScAle (attn)  can lead to  performance similar to  ScAle (both), and thus we mainly report ScAle (attn).  We also report the results with different training sizes in Table~\ref{tab:spatialeval_llava7b}. 20\% of the training data consistently yields the best performance, and we mainly report results using the 20\% training split.


\subsection{Combination with LoRA}
\input{tables/6combine}

While comparing our method against the standard baselines, we notice the significant accuracy gain of   LoRA-all, at the cost of massive trainable parameters compared with ours.  Our ablation study reveals that attention heads make  the most important decision and scaling the output of the attention head provides effective gains in performance over the MLP. 
Inspired by the high accuracy of LoRA and our insights on attention, we combine LoRA and our method with a hybrid setup that retains the LoRA configuration while incorporating ScAle.

Specifically, we apply LoRA to QKV projections in the attention and meanwhile use our method to scale the output  of attention modules. As shown in Table~\ref{tab:spatialeval_lora_kqv_scale_full}, by combining LoRA-QKV and our ScAle, we can achieve an even  higher accuracy on spatial reasoning tasks compared with LoRA-all (such as our 78.9\% v.s. 71.6\% from LoRA-all on Maze-Nav), with only 36\%  trainable parameters. 
It demonstrates that ScAle is a perfect complimentary to LoRA, with better parameter efficiency. Thus ScAle not only provides gain in performance but also gives users a way to determine important network parts and let users design application specific setup along with our light-weight attention head modulation technique. 
As discussed above, ScAle focuses on attention modules with effective improvements and we do not apply LoRA to all modules to save training efforts.

\subsection{Visualizing  Scaling Factors}

\begin{figure}[t]

  \centering
\includegraphics[width=0.9\linewidth]{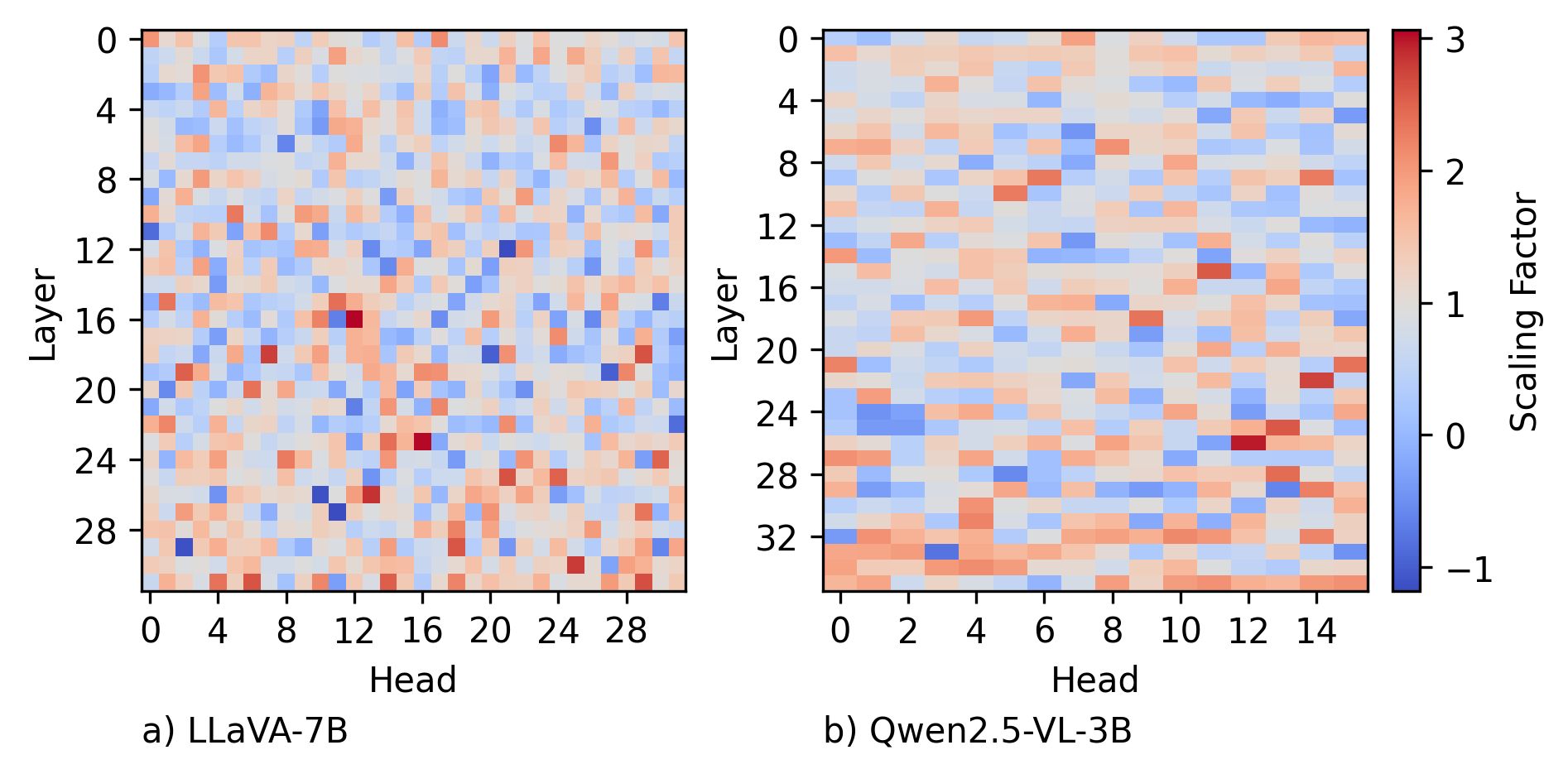}

   \caption{Learned scaling factors for (a) LLaVA-7B and (b) Qwen2.5-VL-3B.}
   \label{fig:attnscalehm}

\end{figure}
The scaling-factor heatmap in Fig.~\ref{fig:attnscalehm} 
reveal a distributed, structured reweighting across heads and layers, learned automatically via gradient descent rather than manual head selection. Such mid-layer emphasis with renewed activation in later layers is common in LLaMA-style transformers \cite{he2024matters}, 
reflecting a transition from mid-depth feature synthesis to late-layer refinement and output calibration.

These findings position bounded activation scaling as a simple and highly parameter-efficient adapter for spatial reasoning in vision--language models.

%% file: tables/1spatial_llava7b.tex
\begin{table}[t]
\centering
\caption{SpatialEval evaluation with LLaVA-7B under different training data size.}
\label{tab:spatialeval_llava7b}
\small
\begin{tabular}{l|l|c|c|c|c|c|c|c|c|c}
\hline

 
Method  &Trainable &\multicolumn{3}{c}{} & \multicolumn{3}{c}{Task} & \multicolumn{3}{c}{} \\
\cline{3-11}
  &Param. &\multicolumn{3}{c|}{Maze-Nav} & \multicolumn{3}{c|}{Spatial-Grid} & \multicolumn{3}{c}{Spatial-Map} \\
\cline{3-11}
 & &5\% & 10\% & 20\% & 5\% & 10\% & 20\% & 5\% & 10\% & 20\% \\
\hline
Frozen& 0& 31.7\% & 31.7\% & 31.7\% & 29.3\% & 29.3\% & 29.3\% & 22.7\% & 22.7\% & 22.7\% \\
(IA)$^{3}$&663,552 & 68.7\% & 70.3\% & 70.8\% & 68.0\% & 79.3\% & 84.8\% & 50.0\% & 59.0\% & 80.7\% \\
LoRA-all&2,646,016& 70.0\% & 71.3\% & 71.6\% & 78.3\% & 80.0\% & 81.9\% & 62.7\% & 79.0\% & 83.2\% \\
LoRA-last &78,080&37.3\% & 46.3\% & 53.6\% & 31.7\% & 30.7\% & 29.2\% & 24.3\% & 27.3\% & 24.3\% \\
\hline
ScAle (attn)&1,024 & 66.7\% & 69.0\% & 68.9\% & 51.7\% & 67.7\% & 68.6\% & 44.3\% & 47.0\% & 45.0\% \\
ScAle (mlp)&32 & 34.0\% & 34.3\% & 44.0\% & 30.0\% & 30.3\% & 30.7\% & 31.7\% & 37.3\% & 38.0\% \\
ScAle (both)&1,056 & 67.7\% & 72.3\% & 69.7\% & 50.7\% & 62.0\% & 75.3\% & 43.0\% & 46.0\% & 45.3\% \\
\hline
\end{tabular}

\end{table}

%% file: tables/2spatial_all.tex
\begin{table}[t]

\centering
\caption{SpatialEval evaluation with different models under 20\% training data.}
\label{tab:spatialeval_best_acc_upto}

\begin{tabular}{l|l|c|c|c|c}
\hline
 & Method & LLaVA-7B & LLaVA-13B & Qwen2.5-VL-3B & Qwen2.5-VL-7B \\
\hline
Params & Frozen & 0 & 0 & 0 & 0 \\
 & (IA)$^3$ & 663.6K & 1.01M & 524.2K & 668.5K \\
 & LoRA-all & 2.65M & 4.06M & 2.32M & 2.97M \\
 & LoRA-last & 78.1K & 97.8K & 52.0K & 90.1K \\
 & ScAle (attn) & 1.0K & 1.6K & 576 & 784 \\
\hline
Maze-Nav & Frozen & 31.7\% & 26.7\% & 26.0\% & 38.3\% \\
 & (IA)$^3$ & 70.8\% & 70.5\% & 82.2\% & 80.9\% \\
 & LoRA-all & 71.6\% & 75.2\% & 79.9\% & 86.2\% \\
 & LoRA-last & 53.6\% & 46.2\% & 41.8\% & 66.5\% \\
 & ScAle (attn) & 68.9\% & 69.8\% & 70.3\% & 66.2\% \\
\hline
Spatial-Grid & Frozen & 29.3\% & 36.7\% & 63.7\% & 73.7\% \\
 & (IA)$^3$ & 84.8\% & 83.9\% & 91.7\% & 91.9\% \\
 & LoRA-all & 81.9\% & 85.2\% & 91.6\% & 92.2\% \\
 & LoRA-last & 29.2\% & 43.5\% & 68.8\% & 83.7\% \\
 & ScAle (attn) & 68.6\% & 62.0\% & 75.8\% & 83.8\% \\
\hline
Spatial-Map & Frozen & 22.7\% & 28.0\% & 51.7\% & 66.3\% \\
 & (IA)$^3$ & 80.7\% & 80.7\% & 88.6\% & 91.6\% \\
 & LoRA-all & 83.2\% & 86.8\% & 91.7\% & 93.4\% \\
 & LoRA-last & 24.3\% & 28.3\% & 46.2\% & 74.1\% \\
 & ScAle (attn) & 45.0\% & 51.5\% & 66.8\% & 76.8\% \\
\hline
\end{tabular}
\end{table}

%% file: tables/3whatup.tex
\begin{table}[t]

\centering
\caption{Test accuracy on WhatsUp-VLM and POPE benchmarks.}
\label{tab:whatsup_last_epoch_rebuttal}

\begin{tabular}{l|l|c|c}
\hline
 & Method & LLaVA-7B & Qwen2.5-VL-3B \\
\hline
Params & (IA)$^3$ & 664.K & 524.K \\
 & ScAle (attn) & 1.02K & 576 \\
\hline
VGQA & Frozen & 56.5\% & 84.5\% \\
 & (IA)$^3$ & 97.8\% & 98.4\% \\
 & ScAle (attn) & 92.9\% & 93.5\% \\
\hline
COCOQA & Frozen & 56.3\% & 83.9\% \\
 & (IA)$^3$ & 96.6\% & 98.4\% \\
 & ScAle (attn) & 91.5\% & 95.0\% \\
\hline
POPE & Frozen & 82.1\% & 87.3\% \\
 & (IA)$^3$ & 89.8\% & 90.4\% \\
 & ScAle (attn) & 88.2\% & 90.4\% \\
 
\hline
\end{tabular}

\end{table}

%% file: tables/5pope.tex

%% file: tables/4cross_task.tex
\begin{table}[t]
\centering
\caption{Cross-task evaluation with Qwen2.5-VL-3B. }
\small
\begin{tabular}{l|c|c|c|c|c|c}
\hline
Evaluated on &Frozen  & \multicolumn{5}{c}{ScAle trained on} \\
\cline{3-7}
 &  & Maze-Nav& Spatial-Grid & Spatial-Map & COCOQA& VGQA \\
\hline
GLUE/SST2& 65.8\% & 58.8\% & 66.1\% & 88.3\% & 54.7\% & 69.0\% \\
GLUE/MRPC& 66.7\% & 74.5\% & 73.8\% & 65.9\% & 69.4\% & 68.4\% \\
Spatial-Map & 51.7\% & 55.9\% & 51.7\% & 66.8\% & 51.5\% & 49.8\% \\
Maze-Nav & 26.0\% & 70.3\% & 27.3\% & 31.3\% & 32.3\% & 28.8\% \\
Spatial-Grid & 63.7\% & 67.0\% & 75.8\% & 65.4\% & 60.4\% & 57.1\% \\
COCOQA & 83.9\% & 91.6\% & 90.9\% & 89.0\% & 95.0\% & 93.5\% \\
VGQA & 84.5\% & 89.9\% & 92.2\% & 88.3\% & 94.7\% & 93.5\% \\
\hline
\end{tabular}

\label{tab:cross_task_grid}
\end{table}

%% file: tables/6combine.tex
\begin{table}[t]

\centering
\caption{SpatialEval accuracy with LLaVA-7B for hybrid design.} 
\label{tab:spatialeval_lora_kqv_scale_full}
\begin{tabular}{l|c|c|c}
\hline
 & Frozen & LoRA\_kqv+ScAle & LoRA-all \\
\hline
Params & 0 & 935 K &  2.65M \\
\hline
Maze-Nav & 31.7\%& 78.9\% & 71.6\% \\
\hline
Spatial-Grid &29.3\% & 85.8\% & 81.9\% \\
\hline
Spatial-Map & 22.7\%  & 85.8\% & 83.2\%  \\
\hline
\end{tabular}
\end{table}

%% file: sec/conclusion.tex
\section{Conclusion}

We introduce \textbf{ScAle}, a minimal adaptation mechanism that learns bounded activation scaling factors to steer frozen vision--language models toward improved spatial reasoning. Across three SpatialEval tasks and four model families, ScAle achieves up to 134.1\% relative accuracy gains over frozen baselines using only $\sim$1K trainable parameters. Beyond synthetic spatial benchmarks, ScAle also improves performance on real-world spatial VQA WhatsUp-VLM and enhances visual grounding robustness on hallucination benchmark. Importantly, these improvements in visual reasoning occur without degrading general language understanding  and often transfer across tasks, indicating that the learned scalars capture transferable spatial structure rather than task-specific heuristics.

Overall, our findings show that lightweight activation modulation can serve as a practical and highly parameter-efficient adapter for vision–language models. This suggests a broader paradigm for adapting large models through minimal, interpretable adjustments rather than heavy parameter updates, offering a promising path for deployment scenarios where bandwidth, storage, and training costs are constrained.
